\documentclass{article}

\usepackage{arxiv}

\usepackage[utf8x]{inputenc} 
\usepackage[T1]{fontenc}    
\usepackage[pagebackref,breaklinks,colorlinks]{hyperref}       
\usepackage{url}            
\usepackage{booktabs}       
\usepackage{amsfonts}       
\usepackage{amsmath}
\usepackage{amssymb}
\usepackage{nicefrac}       
\usepackage{microtype}      
\usepackage[capitalize]{cleveref}       
\usepackage{lipsum}         
\usepackage{graphicx}
\usepackage{natbib}
\usepackage{doi}

\usepackage{caption}

\usepackage{pifont}
\usepackage{arydshln}
\usepackage{multirow}
\usepackage{subcaption}
\usepackage{euscript}
\usepackage{adjustbox}

\newcommand{\refappendix}[1]{\hyperref[#1]{Appendix-\ref*{#1}}}

\newcommand{\cmark}{\ding{51}}%
\crefname{section}{Sec.}{Secs.}
\Crefname{section}{Section}{Sections}
\Crefname{table}{Table}{Tables}
\crefname{table}{Tab.}{Tabs.}

\title{PCT: Perspective Cue Training Framework\\for Multi-Camera BEV Segmentation}

\date{March 18, 2024}

\usepackage{authblk}

\newbox{\orcid}\sbox{\orcid}{\includegraphics[scale=0.06]{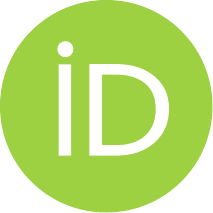}} 
\author[1,*]{%
  Haruya Ishikawa
}
\author[2]{%
	Takumi Iida
}
\author[2]{%
  Yoshinori Konishi
}
\author[1]{%
  Yoshimitsu Aoki
}
\affil[1]{Department of Electrical Engineering, Keio University, Kanagawa, Japan}
\affil[2]{SenseTime Japan, Kyoto, Japan}
\affil[*]{This work was done while the author was an intern at SenseTime Japan.}

\renewcommand{\shorttitle}{PCT: Perspective Cue Training Framework for Multi-Camera BEV Segmentation}

\hypersetup{
pdftitle={PCT: Perspective Cue Training Framework for Multi-Camera BEV Segmentation},
pdfsubject={},
pdfauthor={Haruya Ishikawa, Takumi Iida, Yoshinori Konishi, Yoshimitsu Aoki},
pdfkeywords={BEV Segmentation, Semi-Supervised Learning, Unsupervised Domain Adaptation, Computer Vision},
}

\begin{document}

\maketitle

\begin{abstract}
Generating annotations for bird's-eye-view (BEV) segmentation presents significant challenges due to the scenes' complexity and the high manual annotation cost.
In this work, we address these challenges by leveraging the abundance of unlabeled data available.
We propose the Perspective Cue Training (PCT) framework, a novel training framework that utilizes pseudo-labels generated from unlabeled perspective images using publicly available semantic segmentation models trained on large street-view datasets.
PCT applies a perspective view task head to the image encoder shared with the BEV segmentation head, effectively utilizing the unlabeled data to be trained with the generated pseudo-labels.
Since image encoders are present in nearly all camera-based BEV segmentation architectures, PCT is flexible and applicable to various existing BEV architectures.
PCT can be applied to various settings where unlabeled data is available.
In this paper, we applied PCT for semi-supervised learning (SSL) and unsupervised domain adaptation (UDA).
Additionally, we introduce strong input perturbation through Camera Dropout (CamDrop) and feature perturbation via BEV Feature Dropout (BFD), which are crucial for enhancing SSL capabilities using our teacher-student framework.
Our comprehensive approach is simple and flexible but yields significant improvements over various baselines for SSL and UDA, achieving competitive performances even against the current state-of-the-art. 
\end{abstract}


\section{Introduction}
\label{sec:intro}

Bird's-eye-view (BEV) segmentation is crucial for autonomous driving and navigation systems, providing a comprehensive spatial understanding of the vehicle's surroundings.
However, creating accurate BEV annotation is notoriously tricky and resource-intensive, requiring sensor calibrations and accurate 3D annotations.
Additionally, the performance of BEV segmentation models often suffers from domain shifts, where models trained in one environment underperform in another due to differences in geography, weather, and lighting conditions.
Semi-supervised learning (SSL) and unsupervised domain adaptation (UDA) emerge as promising approaches to mitigate these challenges by effectively utilizing unlabeled data.

SSL and UDA are paradigms that aim to improve model performance by utilizing labeled and unlabeled data sets during training.
In SSL, the unlabeled data typically come from the same distribution as the labeled data, allowing the model to learn more generalizable features.
On the other hand, UDA deals with scenarios where the unlabeled data exhibit characteristics of domain gaps, such as variations in weather conditions, lighting, or urban landscapes.
For instance, an autonomous vehicle trained in one city may need to adapt to different weather conditions, architectural styles, and even traffic regulations, such as diving on opposite sides of the road, when deployed in another city.

Recent progress in annotation approaches like VMA \cite{Chen2023VMA} and CAMA \cite{Zhang2023CAMA} has shown promise in reducing the cost of manual annotation for HD maps.
However, these methods are still prone to errors and annotating all possible scenarios and semantic categories is still infeasible.

\begin{figure}[ht]
\centering
\begin{minipage}[t]{0.45\textwidth}
\centering
\includegraphics[width=\textwidth]{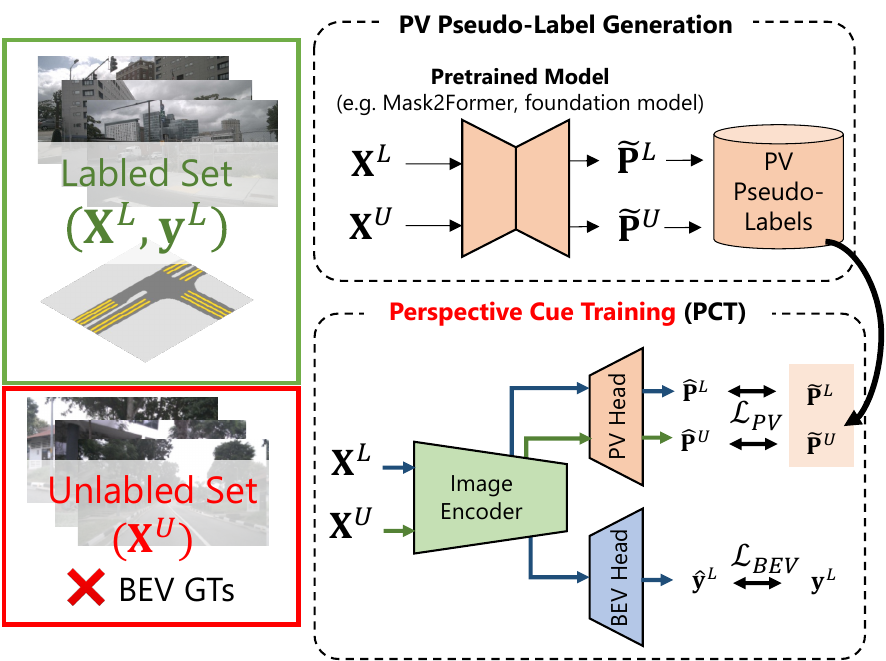}
\end{minipage}
\vfill
\begin{minipage}[t]{0.45\textwidth}
\centering
\includegraphics[width=\textwidth]{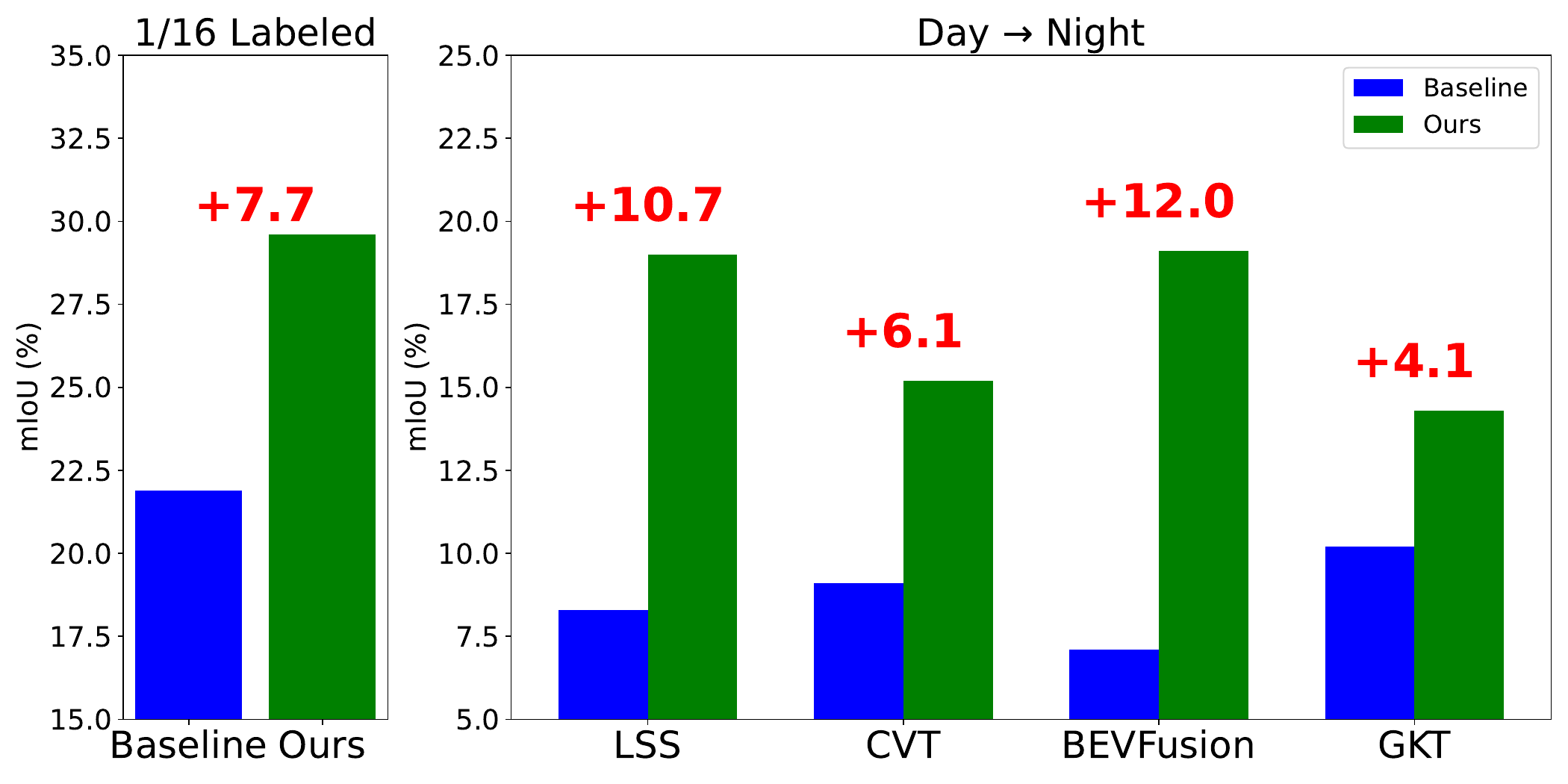}
\end{minipage}
\caption{\textbf{General overview of our proposed Perspective Cue Training (PCT) framework and the impact it has on tasks requiring the utilization of unlabeled data.} PCT framework utilizes PV pseudo-labels generated from easily accessible models (e.g. Mask2Former for semantic segmentation) to train multi-camera BEV segmentation models. PCT is flexible and applicable to various BEV architectures. Our method significantly improves the baseline for both SSL and UDA tasks utilizing unlabeled data.}
\label{fig:overview}
\end{figure}

In this paper, we propose the Perspective Cue Training (PCT) framework, which utilizes perspective view (PV) tasks for settings where an abundance of unlabeled data are available, like SSL and UDA.
We noticed that abundant PV images, taken from the vehicle's cameras, are unlabeled for datasets related to BEV perception \cite{Caesar2019nuScenes, Wilson2023Argoverse2}.
The recent surge in the performance of PV tasks, notably semantic segmentation, has been remarkable, thanks to publicly available pretrained models and datasets like Cityscapes and BDD100k \cite{Cordts2016Cityscapes, Yu2020BDD100k, Ros2016SYNTHIA, Chen2018DLV3, Xie2021SegFormer, Cheng2021Mask2Former, mmseg2020}.
These advancements allow us to generate pseudo-labels for many unlabeled PV images used in BEV segmentation training.
The PCT framework capitalizes on this by training the image encoder with an additional PV head in a multi-task learning manner with the BEV segmentation head, effectively utilizing the pseudo-labels generated from the PV tasks.
PCT is flexible because image encoders exist in nearly all camera-based BEV segmentation architectures, making it applicable to various existing BEV architectures.
Furthermore, there are no added computational costs during inference because the PV head is only required during training.
Utilizing PV pseudo-labels is an easy method to boost scenarios where BEV annotations are scarce, such as SSL and UDA, as shown in Figure \ref{fig:overview}.
Note that PCT only generates PV pseudo-labels, not BEV pseudo-labels.

Furthermore, we introduce additional modules such as Camera Dropout (CamDrop) and BEV Feature Dropout (BFD) to refine our training strategy further.
These modules, particularly when combined with PCT, demonstrate vital performance improvements for SSL and UDA.

The contributions of this work are as follows:
\begin{itemize}
\item We tackle the challenge in BEV segmentation, where labeled data is limited, by leveraging PV images of labeled and unlabeled data with our proposed Perspective Cue Training (PCT) framework. We demonstrate the effectiveness of our approach in both SSL, where BEV annotations in the source domain are scarce, and UDA, where BEV annotations in the target domain are unavailable.
\item To our knowledge, SSL for multi-camera BEV segmentation has not been previously explored. We propose baselines based on SSL methods from other tasks like semantic segmentation. We introduce techniques such as CamDrop and BFD to further enhance the SSL capabilities of not just our method but can also be applied to the baseline methods.
\item For UDA, we compare our method against various baselines and the current state-of-the-art approach, DualCross \cite{Man2023DualCross}, which utilizes a teacher model trained on LiDAR data. Our method shows superior performance in most benchmarks without relying on additional modalities.
\end{itemize}

\section{Related Works}
\label{sec:related_works}

\noindent\textbf{Camera-based BEV Segmentation.}
Recent advances in camera-based BEV segmentation have leveraged multi-camera setups to reconstruct the 3D scene and project it onto a BEV plane.
Notable methods include LSS \cite{Philion2020LSS}, BEVDepth \cite{Li2022BEVDepth}, BEVFusion \cite{Liu2022BEVFusion}, CVT \cite{Zhou2022CVT}, and GKT \cite{Chen2022GKT}. 
LSS encodes images from arbitrary camera rigs by implicitly unprojecting them to 3D, offering a flexible framework for BEV segmentation.
BEVDepth extends LSS by leveraging explicit depth supervision of the 2D-to-BEV module.
BEVFusion fuses multi-task and multi-sensor data with a unified BEV representation, showcasing the benefits of integrating diverse data sources for improved segmentation and obtaining SOTA performance on the nuScenes dataset \cite{Caesar2019nuScenes}.
CVT proposes a novel transformer-based 2D-to-BEV module that implicitly learns to map individual camera views into BEV representation.
Finally, GKT improves upon CVT by introducing efficient and robust 2D-to-BEV representation learning with geometric cues to enhance the transformation process.

Unique methods of data augmentation to increase robustness for BEV perception has been proposed.
\cite{Klinghoffer2023TowardsVR} proposes a novel view synthesis approach to generate view points of different sensor rigs for monocular BEV perception.
BEVGen \cite{Swerdlow2023BEVGen} introduces a method to generate multi-camera views from BEV segmentation map as a way to augment the dataset.

Additionally, the integration of PV cues has shown promise in augmenting BEV segmentation.
X-Align \cite{Borse2022XAlign} demonstrates that cross-modal cross-view alignment, facilitated by PV multi-task learning with PV pseudo-labels, can significantly boost BEV segmentation performances.
While our work shares similarities with X-Align in utilizing PV pseudo-labels, we extend this concept by employing a PV task head to leverage unlabeled PV images.
This approach is effective in scenarios with limited BEV annotations, as we show in our SSL and UDA experiments.

\noindent\textbf{Semi-Supervised Learning (SSL).}
SSL is essential for tasks like BEV segmentation where labeled data is limited.
Techniques like pseudo-labeling and consistency regularization have been explored, with the Mean-Teacher framework \cite{Tarvainen2017MeanTeacher} being notable for its effectiveness across various tasks \cite{Ke2019DualSB, Chen2020AMM, Xu2021SoftTeacher}.
Input perturbations, like Cutout \cite{Devries2017Cutout} and CutMix \cite{Yun2019CutMix}, are crucial for SSL in semantic segmentation \cite{French2019SemisupervisedSS}.
UniMatch \cite{Yang2022UniMatch}, based on FixMatch \cite{Sohn2020FixMatch}, and CCT \cite{Ouali2020CCT} utilizes feature perturbations to enhance SSL capabilities. 

In BEV segmentation, SSL is less explored.
SkyEye \cite{Gosala2023SkyEye} uses a self-supervised approach of effectively utilizing BEV pseudo-labels generated from PV semantic segmentation task, while \cite{Zhu2023SSLMonoBEV} introduces conjoint rotation for augmentation to increase supervised data.
Both of these methods are for monocular BEV segmentation and no literature exists on multi-camera BEV segmentation to the best of our knowledge.
Our work introduces a PV sub-task to leverage unlabeled data for SSL.
We also introduce CamDrop and BFD for input and feature perturbations, respectively.
These enhancements are vital for applying SSL frameworks like Mean-Teacher and UniMatch to BEV segmentation.

\noindent\textbf{Unsupervised Domain Adaptation (UDA).}
UDA aims to adapt models trained on a labeled source domain to perform well on an unlabeled target domain.
Techniques such as Adversarial Discriminative Domain Adaptation \cite{Tzeng2017ADDA}, Maximum Classifier Discrepancy \cite{Saito2017MaximumCD}, and Deep Adaptation Networks \cite{Long2015LearningTF} have been proposed to minimize domain discrepancies.
In the context of pixel-wise classification methods, traditional UDA methods like Fourier Domain Adaptation (FDA) \cite{Yang2020FDA} and domain adversarial training with Gradient Reversal Layers (GRL) \cite{Ganin2015DomainAdversarial} along with contrastive learning provide a foundation for recent UDA methods \cite{Hoyer2021DAFormer, Zhao2023UnsupervisedDA}.

UDA for BEV domain is under-explored, with DualCross \cite{Man2023DualCross} and DA-BEV \cite{Jiang2024DABEV} being the only methods.
DualCross employs a knowledge distillation scheme and a two-stage training process, leveraging cross-modal knowledge from LiDAR.
DA-BEV, on the other hand, applies self-training with features from both the image and BEV spaces, along with adversarial learning in both spaces.
We focus on a flexible camera-only framework that can be applied to various existing architectures by leveraging pseudo-labels from unlabeled PV images.


\section{Approach}
\label{sec:approach}

This work addresses the task of BEV segmentation of street-view scenes from multi-camera rigs, focusing on leveraging the abundance of unlabeled data to enhance model performance.
We approach this through the paradigms of Semi-Supervised Learning (SSL) and Unsupervised Domain Adaptation (UDA), utilizing a combination of labeled and unlabeled datasets during training.
As explained before, SSL and UDA aim to improve model performance by utilizing both labeled and unlabeled data sets during training.

For coherent methodology, we unify the two tasks, where we have a labeled set $\mathbb{L}$ and an unlabeled set $\mathbb{U}$.
For SSL, the labeled set $\mathbb{L}$ and the unlabeled set $\mathbb{U}$ are from the same domain, with the number of labeled samples being much smaller than the number of unlabeled samples, i.e., $\lvert \mathbb{L} \rvert \ll \lvert \mathbb{U} \rvert$.
In UDA, the labeled set $\mathbb{L}$ is from the source domain, while the unlabeled set $\mathbb{U}$ is from the target domain, with a domain shift between the two.
The labeled set contains a set of multi-camera data $\mathbf{X}^L$ and BEV ground truth (GT) $\mathbf{y}^L$ where $(\mathbf{X}^L, \mathbf{y}^L) \in \mathbb{L}$, while the unlabeled set only contains multi-camera data $\mathbf{X}^U \in \mathbb{U}$.

We define $\mathbf{X}=\{\mathbf{X}_i\}^N_{i=1}=\{\mathbf{I}_i, \mathbf{K}_i, \mathbf{R}_i, \mathbf{t}_i\}^{N}_{i=1}$ as a batch of multi-view images, where $\mathbf{I}_i$ is the image, $\mathbf{K}_i$ is the intrinsic parameter, $\mathbf{R}_i$ and $\mathbf{t}_i$ are the extrinsic for the $i$-th camera in the batch, and $N$ is the number of cameras on the vehicle.
Following \cite{Liu2022BEVFusion}, we treat the BEV segmentation task as multi-label classification where each pixel can have multiple labels.
Therefore, BEV GT is formalized as $\mathbf{y}^L \in \{0, 1\}^{H\times W\times C}$, where $H$ and $W$ are the height and width of the BEV grid, and $C$ is the number of classes.

Multi-camera BEV segmentation models typically comprise the following components: an Image Encoder, a 2D-to-BEV Module, and a BEV Encoder and Decoder, with the BEV Encoder sometimes omitted.
Methods like CVT \cite{Zhou2022CVT} have an additional BEV embedding as an input, but we omit these method-specific components in our discussion for brevity.
The processing flow of these modules is as follows:
\begin{align}
  \mathbf{F}_{image} &= \operatorname{Image Encoder}(\mathbf{I}) \\
  \mathbf{f}_{bev} &= \operatorname{2DtoBEV}(\mathbf{F}_{image}, \mathbf{K}, \mathbf{R}, \mathbf{t}) \\
  \mathbf{f}'_{bev} &= \operatorname{BEV Encoder}(\mathbf{f}_{bev}) \\
  \mathbf{\hat{y}} &= \operatorname{BEV Decoder}(\mathbf{f}'_{bev}), \label{equ:pred}
\end{align}
where $\mathbf{F}_{image}$ is the image features obtained from multi-camera images, $\mathbf{f}_{bev}$ is the BEV feature, and $\mathbf{\hat{y}}$ is the output BEV prediction.

For brevity, we simplify our notation by treating $\mathbf{X} = \mathbf{I}$ in subsequent sections and assume that the correct camera parameters are provided to the $\operatorname{2DtoBEV}$ module.

Our method is organized as follows:
\begin{itemize}
  \item Section \ref{sec:pvc} introduces the Perspective Cue Training (PCT) framework, which leverages pseudo-labels of perspective view task to utilize unlabeled data for BEV segmentation.
  \item Section \ref{sec:camdrop} presents Camera Dropout (CamDrop), a novel input perturbation technique for multi-camera BEV segmentation.
  \item Our training strategy with PCT for UDA is formalized in Section \ref{sec:uda_training}.
  \item Finally, Section \ref{sec:semisup_training} details our SSL approach, which incorporates BEV Feature Dropout (BFD) and a teacher-student network training strategy.
\end{itemize}

\subsection{Perspective Cue Training (PCT) Framework}
\label{sec:pvc}

\begin{figure}[ht]
   \centering
   \includegraphics[width=0.4\linewidth]{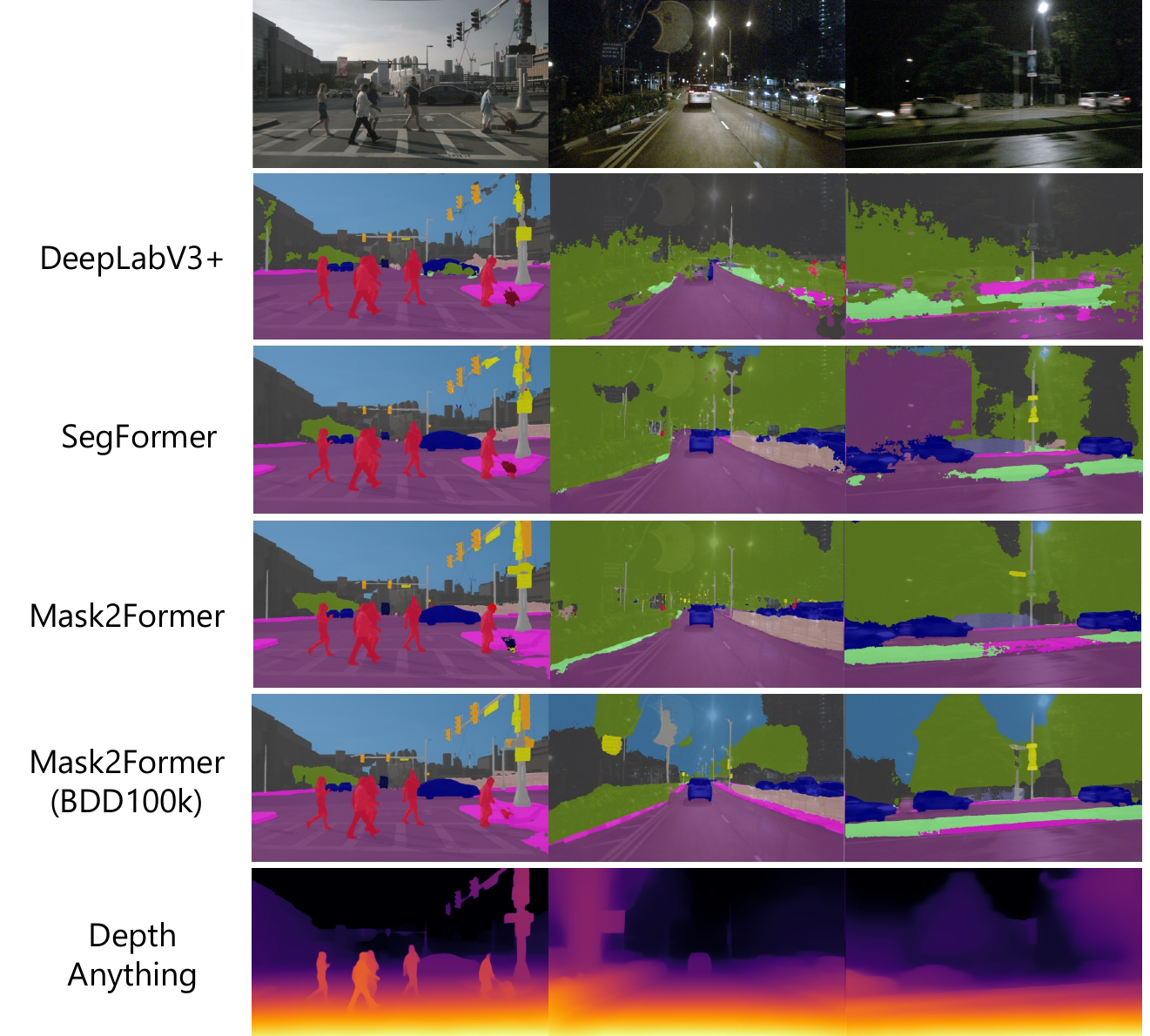}
   \caption{\textbf{Visualization of pseudo-labels generated by easily accessible models on the nuScenes dataset.} We generate semantic segmentation pseudo-labels from pretrained models; namely DeepLabV3+ \cite{Chen2018DLV3}, SegFormer \cite{Xie2021SegFormer}, and Mask2Former \cite{Cheng2021Mask2Former}. Out of the predictions trained on Cityscapes \cite{Cordts2016Cityscapes} (rows 2 to 4), Mask2Former exhibits the cleanest results especially for harder domains like nighttime. When trained on BDD100k \cite{Yu2020BDD100k}, which has diverse scenarios such as weather and time of day, the pseudo-label are qualitatively better, where over- and under-segmentation occurs less frequently. We also explore the use of relative depth pseudo-labels obtained from Depth Anything \cite{Yang2024DepthAnything}. Best viewed in color and zoomed in.}
   \label{fig:pseudo_label_quality}
\end{figure}

We propose the Perspective Cue Training (PCT) framework, which utilizes the PV pseudo-labels to train the BEV segmentation model in multi-task learning manner.
This framework is only applied during the training stage of the target BEV segmentation model.
In our work we first generate pseudo-labels of perspective images from all sets $\mathbf{X}^{L \cup U}$, including both labeled $\mathbf{X}^L$ and unlabeled images $\mathbf{X}^U$, where $\mathbf{X}^{L \cup U} \in (\mathbb{L} \cup \mathbb{U})$.
We chose semantic segmentation for our main pseudo-labeling task because the pixel-wise classification task is similar to BEV segmentation.
However, we have also experimented with relative depth estimation task, as shown in Figure \ref{fig:pseudo_label_quality}.
We generally use Mask2Former \cite{Cheng2021Mask2Former} trained on BDD100k \cite{Yu2020BDD100k} as our default pseudo-label generator, but we explore the effects of other network architectures and training datasets in our ablation studies.
Our pseudo-label generator $\operatorname{PLGen}$ obtains $\tilde{\mathbf{P}} = \{\operatorname{PLGen}(\mathbf{x}_i) \mid \forall \mathbf{x}_i \in \mathbf{X}^{L \cup U}\}$, where $\tilde{\mathbf{P}}$ are one-hot encoded pseudo-labels.

PV task head is applied to the image encoder, and the entire BEV segmentation model is trained in a multi-task learning manner, as shown in Figure \ref{fig:overview}.
We utilize FPN with UPerNet \cite{Xiao2018UPerNet} as our PV task head ($\operatorname{PVHead}$) in the belief that utilizing all the hierarchical features of the image encoder is essential to condition the shared image encoder with PV cues.
We can obtain PV predictions $\hat{\mathbf{P}}$ from the PV task head $\operatorname{PVHead}$ as follows:
\begin{align}
  \mathbf{F}^L_{image} &= \operatorname{ImageEncoder}(\mathbf{X}^L) \\
  \mathbf{F}^U_{image} &= \operatorname{ImageEncoder}(\mathbf{X}^U) \\
  \hat{\mathbf{P}} &= \{\operatorname{PVHead}(\mathbf{f}_i) \mid \forall \mathbf{f}_i \in [\mathbf{F}^L_{image}; \mathbf{F}^U_{image}] \}.
\end{align}

PV loss $\EuScript{L}_{PV}$ is computed using cross-entropy loss between the prediction $\hat{\mathbf{P}}$ and the pseudo-labels $\tilde{\mathbf{P}}$ as follows:
\begin{equation}
\EuScript{L}_{PV} = \frac{1}{N}\sum_{i=1}^{N} \operatorname{CE}(\hat{\mathbf{p}}_i, \tilde{\mathbf{p}}_i),
\end{equation}
where $\hat{\mathbf{p}}_i \in \hat{\mathbf{P}}$ and $\tilde{\mathbf{p}}_i \in \tilde{\mathbf{P}}$ are the $i$-th output probability map of the PV prediction and one-hot encoded pseudo-labels, respectively.

The final multi-task training loss is as follows:
\begin{equation}\label{equ:pv_mtl_loss}
  \EuScript{L}_{PCT} = \EuScript{L}_{BEV} + \lambda_{PV} \EuScript{L}_{PV},
\end{equation}
where $\EuScript{L}_{BEV} = \operatorname{FocalLoss}(\hat{\mathbf{y}}^L, \mathbf{y}^L)$ and $\lambda_{PV}$ is the weight for the PV loss.
Note that $\hat{\mathbf{y}}^L$ is obtained from Equation \ref{equ:pred}.

\subsection{Camera Dropout (CamDrop) Augmentation}
\label{sec:camdrop}

\begin{figure}[ht]
  \centering
  \includegraphics[width=0.4\linewidth]{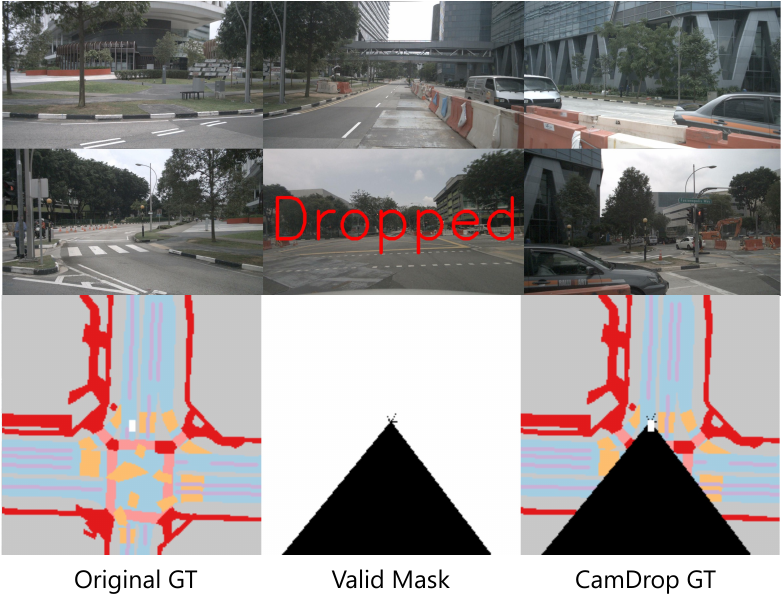}
  \caption{\textbf{Visualization showcasing how Camera Dropout (CamDrop) augmentation is applied to perspective views and BEV ground truth (GT).} Back-viewing camera out of the six cameras is dropped and subsequent areas of the BEV GT, only visible by the dropped camera, are masked out.}
  \label{fig:camdrop}
\end{figure}

Applying traditional pixel-wise input augmentations to BEV segmentation is challenging due to its 3D nature.
For instance, masking sections of the PV image, as in Cutout, would require corresponding masking in the BEV ground truth, which is not straightforward.
To address this, we introduce Camera Dropout (CamDrop), a simple yet effective input perturbation inspired by Cutout \cite{Devries2017Cutout}.
CamDrop randomly drops cameras and masks the associated visible areas in the BEV ground truth, as illustrated in Figure \ref{fig:camdrop}.
This augmentation is efficient, as the camera parameters can easily determine the horizontal viewport.
The perspective view image and the masked BEV area are labeled with ignore labels, ensuring that the dropped regions do not contribute to the loss.
Importantly, we only mask regions exclusively visible from the dropped cameras, ensuring that the masked areas are not visible from the remaining cameras.

\subsection{Training Method for UDA}
\label{sec:uda_training}

The PCT framework provides a method of training on both domains using a shared image encoder, which motivates the model to learn domain-invariant features through PV tasks.
The intuition can be explained using the theorem proposed in \cite{BenDavid2010UDA}, which states that the upper-bound of the target domain error is composed of the source domain error and the domain discrepancy where the latter can be minimized through utilizing both domains with $\EuScript{L}_{PV}$.

For UDA, we formalize labeled set $\mathbb{L}$ as the source domain set and unlabeled set $\mathbb{U}$ as the target domain set.
The loss for training UDA is as follows:
\begin{equation}\label{equ:uda_loss}
  \EuScript{L}_{UDA} = \EuScript{L}_{BEV} + \lambda_{PV} \EuScript{L}_{PV},
\end{equation}
where we use the same loss as in Equation \ref{equ:pv_mtl_loss}.

Additionally, we utilize CamDrop to further enhance the model's robustness.

\subsection{Training Method for SSL}
\label{sec:semisup_training}

\begin{figure}[ht]
  \centering
  \includegraphics[width=0.8\linewidth]{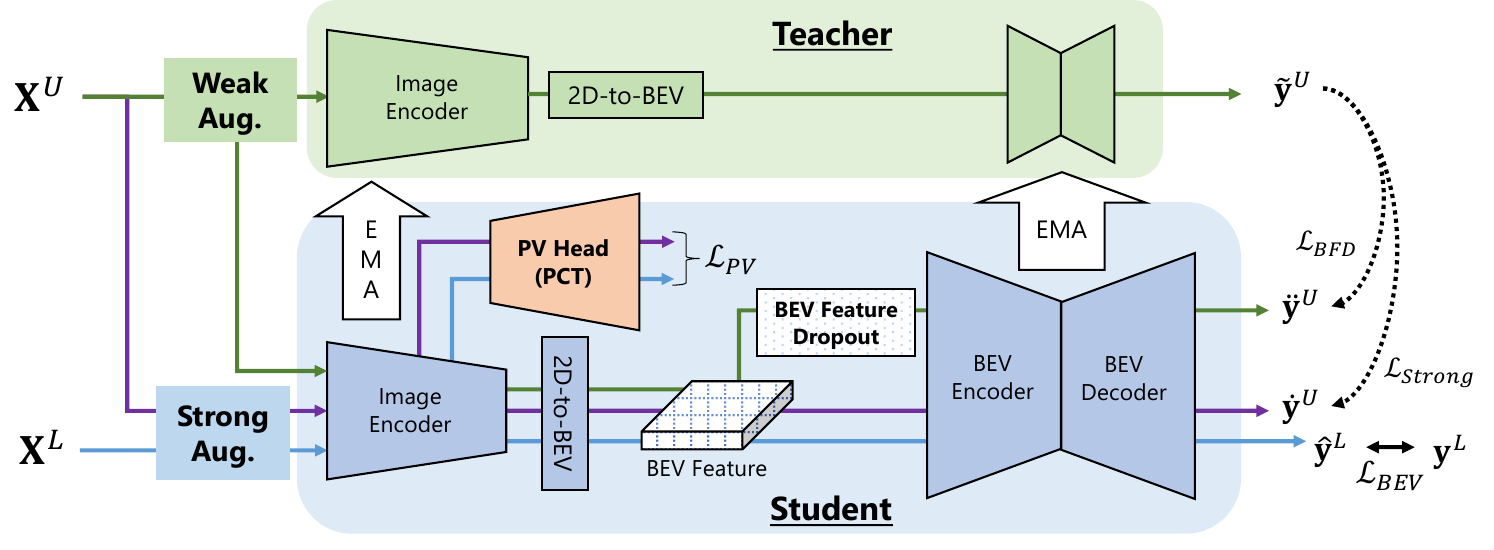}
  \caption{\textbf{Our proposed semi-supervised learning (SSL) training framework utilizing the proposed PCT, Camera Dropout (CamDrop) augmentation, and BEV Feature Dropout (BFD).} The BEV segmentation model jointly trained with pseudo-labels obtained from the perspective view task models using PCT. We utilize the mean-teacher (MT) framework to effectively use the proposed input and feature perturbations by enforcing consistency with the teacher model.}
  \label{fig:semisup_model}
\end{figure}

For SSL, we introduce a teacher-student training framework following the Mean-Teacher (MT) framework \cite{Tarvainen2017MeanTeacher}, as shown in Figure \ref{fig:semisup_model}.
In this framework, we have two identical models, the student and the teacher, where the teacher model is an exponential moving average (EMA) of the student model, computed as:
\begin{equation}
  \theta'_{teacher} = \alpha \theta_{teacher} + (1 - \alpha) \theta_{student},
\end{equation}
where $\theta$ is the model parameters and $\alpha$ is the momentum.

For the teacher network, a weakly augmented input is used, while a strongly augmented input is passed to the student network.
The consistency loss is computed between the weakly augmented teacher's prediction and the strongly augmented student's prediction:
\begin{align}
  \tilde{\mathbf{y}}^U &= \operatorname{Teacher}(\operatorname{WeakAug}(\mathbf{X}^U)) \\
  \dot{\mathbf{y}}^U &= \operatorname{Student}(\operatorname{StrongAug}(\mathbf{X}^U)) \\
  \EuScript{L}_{Strong} &= \frac{1}{M}\sum^{M}_{i=1}(\tilde{y}_i^U - \dot{y}_i^U)^2,
\end{align}
where we calculate the mean squared error for all $M=H\times W \times C$ pixels of the BEV grid.
For $\operatorname{WeakAug}$, we employ augmentations commonly used in BEV segmentation: random horizontal flip, rotation, scaling, and cropping.
For $\operatorname{StrongAug}$, we apply the same augmentations as $\operatorname{WeakAug}$ but with ColorJitter, GaussianBlur, and CamDrop.

We further propose a feature perturbation called BEV Feature Dropout (BFD) inspired by UniMatch \cite{Yang2022UniMatch}.
We apply Dropout \cite{Srivastava2014Dropout} of $50\%$ to the BEV feature maps to obtain perturbed BEV features.
The consistency loss is then computed between the weakly augmented teacher's predictions and the perturbed prediction:
\begin{align}
  \mathbf{f}^U_{BEV} &= \operatorname{2DtoBEV}(\operatorname{ImageEncoder}(\operatorname{WeakAug}(\mathbf{X}^U))) \\
  \ddot{\mathbf{y}}^U &= \operatorname{BEVDecoder}(\operatorname{BEVEncoder}(\operatorname{BFD}(\mathbf{f}^U_{BEV}))) \\
  \EuScript{L}_{BFD} &= \frac{1}{M}\sum^{M}_{i=1}(\tilde{y}_i^U - \ddot{y}_i^U)^2.
\end{align}

The final loss used for SSL is as follows:
\begin{equation}\label{equ:ssl_loss}
  \EuScript{L}_{SSL} = \EuScript{L}_{BEV} + \lambda_{PV} \EuScript{L}_{PV} + \lambda_{Strong} \EuScript{L}_{Strong} + \lambda_{BFD} \EuScript{L}_{BFD},
\end{equation}
where $\lambda_{Strong}$ and $\lambda_{BFD}$ are the weights for strong consistency loss and the BFD consistency loss, respectively.

We note that BFD's benefits are primarily realized through MT framework, which empirically showed strong performance in SSL, but not for UDA.

\section{Experiments}
\label{sec:experiments}

\begin{figure}[h]
   \centering
   \includegraphics[width=0.90\textwidth]{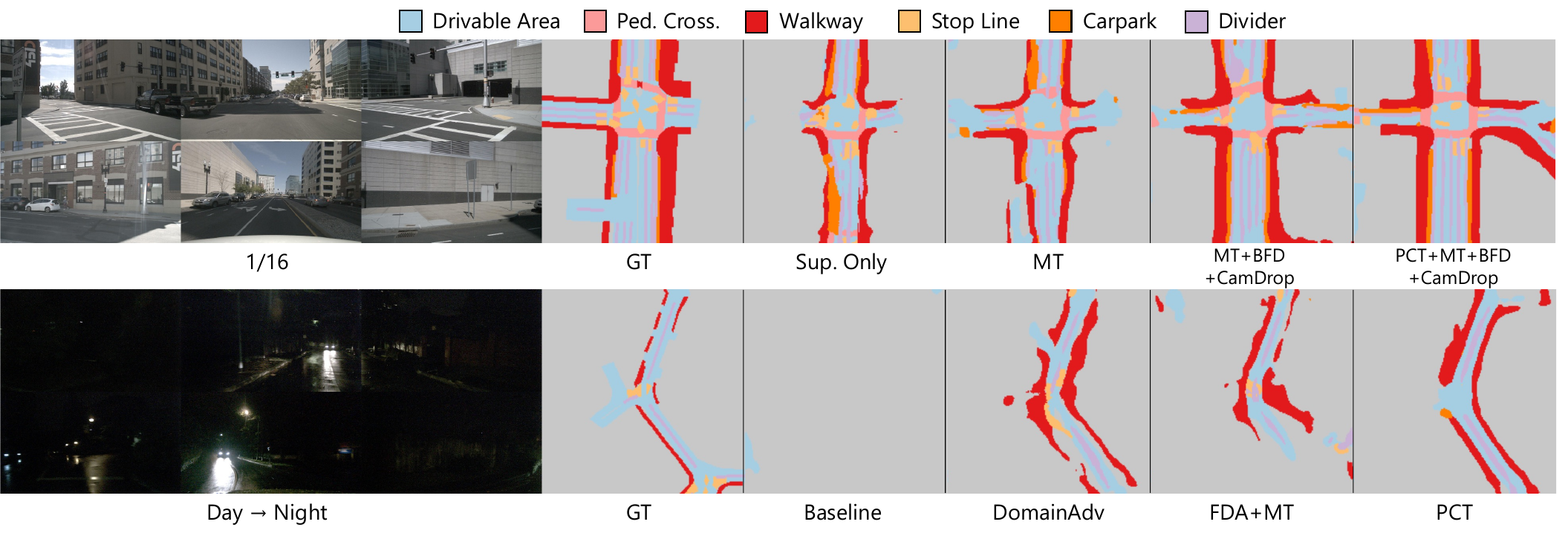}
   \caption{\textbf{Qualitative results for semi-supervised learning (SSL) on the 1/16 split and unsupervised domain adaptation (UDA) on the ``Day $\rightarrow$ Night'' split. Best viewed in color and zoomed in.}}
   \label{fig:qualitative}
\end{figure}

\subsection{Experimental Setup}
\noindent\textbf{SSL Experimental Setup.}
We generate four SSL splits for 1/16, 1/8, 1/4, and 1/2 labeled data from the nuScenes dataset's \cite{Caesar2019nuScenes} training set and treat the remaining labeled data as unlabeled data.
Note that we divide the training scenes into labeled and unlabeled data, not individual sample frames.
For testing, we evaluate on the validation scenes, which is the same for all SSL splits.

\noindent\textbf{UDA Experimental Setup.}
We follow the splits introduced by DualCross \cite{Man2023DualCross} and use the following domain gaps: Day $\rightarrow$ Night, Dry $\rightarrow$ Rain, and Boston $\rightarrow$ Singapore for the nuScenes dataset.
We added Singapore $\rightarrow$ Boston domain gap since Boston $\rightarrow$ Singapore contains mixed domain gaps, as explained in \cite{Man2023DualCross}.

\noindent\textbf{Common Experimental Setup.}
Following \cite{Liu2022BEVFusion}, we segment static categories, which includes the following: Drivable Area, Pedestrian Crossing, Walkway, Stop Line, Carpark Area, and Divider.
We measure the performance using mean Intersection over Union (mIoU).

\noindent\textbf{Implementation Details.}
We base our BEV segmentation code base on \cite{Liu2022BEVFusion} and mmsegmentation \cite{mmseg2020}.
The hyperparameters are all consistent to ensure fair comparison across different methods.
Unless explicitly stated, the crop size is $224\times480$, batch size is $32$, total training iteration is $30k$, the optimizer is AdamW with a learning rate of $0.004$ and weight decay of $0.01$, and the learning rate is scheduled using OneCycle Learning Rate Scheduler.
For all experiments, we present the metric of the final checkpoint.
The baseline method of our work is LSS \cite{Philion2020LSS} with EfficientNet-b4 backbone.
For loss weights in Equations \ref{equ:uda_loss} and \ref{equ:ssl_loss}, we set $\lambda_{PV}=0.1$, $\lambda_{Strong}=0.1$, and $\lambda_{BFD}=0.5$.
As stated in Section \ref{sec:semisup_training}, the weak augmentations are random horizontal flip, rotation, scaling, and cropping, while the default strong augmentations are the same as the weak augmentations but with ColorJitter and GaussianBlur.
For UDA, we use the strong augmentations without GaussianBlur.
The baseline methods use strong augmentations by default unless explicitly stated.
The momentum for the teacher model is set to $\alpha=0.999$.
Based on the original Mean-Teacher implementation, we utilize a sigmoid rampup function for the consistency loss weight, which starts at $0$ and gradually increases to $1$ over the first $9k$ iterations.
We train and validate all of our experiments using 8 NVIDIA V100 GPUs.

\noindent\textbf{Pseudo-label Generation.}
The nuScenes dataset does not have semantic segmentation annotations for the PV images.
We utilized mmsegmentation \cite{mmseg2020} to generate PV pseudo-labels.
We retrained Mask2Former with the same configuration as the Cityscapes dataset on datasets without pretrained weights.
For relative depth estimation, we use the publicly available code provided by Depth Anything \cite{Yang2024DepthAnything}.

This task can be extended further by allowing the target image to be anything.
For example, instead of an image, we could potentially train the agent to find a specific object or person in a scene by looking around.
This would increase the motivation of searching and tracking people using PTZ like cameras which are quite common for CCTV in public environments.
While the application could increase security and safety, it can be abused by authorities and cause privacy infringement in certain situations.

\subsection{Semi-Supervised Learning Results}
\label{sec:ssl_results}

\begin{table}[h]
\centering
\caption{Semi-Supervised Learning results on the nuScenes dataset. We experimented on four levels of labeled data and evaluated on the validation set. ``CamDrop'' is our proposed Camera Dropout and ``BFD'' is the BEV Feature Dropout. The results are in mIoU (\%) and the best results are in \textbf{bold}.}
\label{tab:ssl_nusc}
\footnotesize
\begin{tabular}{c|cc||c|c|c|c}
\hline
Method    & CamDrop & BFD    & 1/16 & 1/8  & 1/4  & 1/2  \\
\hline
Sup. Only               &        &        & 21.9 & 27.4 & 36.4 & 47.0 \\
\hdashline
\multirow{4}{*}{MT}     &        &        & 23.7 & 29.8 & 37.5 & 47.9 \\
                        & \cmark &        & 25.5 & 31.4 & 38.6 & 47.5 \\
                        &        & \cmark & 25.2 & 31.8 & 39.3 & 49.1 \\
                        & \cmark & \cmark & 27.0 & 32.6 & 40.1 & 49.1 \\
\hdashline
UniMatch                & \cmark & \cmark & 22.9 & 29.3 & 39.2 & 49.0 \\
\hdashline
\multirow{2}{*}{PCT}   &        &        & 24.1 & 29.3 & 37.6 & 48.7 \\
                       & \cmark &        & 24.9 & 30.0 & 38.3 & 48.4 \\
\hdashline
\multirow{2}{*}{PCT+MT}& \cmark &        & 28.6 & 34.0 & 40.7 & 50.4 \\
                       & \cmark & \cmark & \textbf{29.6} & \textbf{35.0} & \textbf{41.9} & \textbf{51.6} \\
\hline
\end{tabular}
\end{table}

In our SSL benchmark, we compare our proposed method against the following baselines:
\begin{itemize}
  \item \textbf{Sup. Only}: LSS supervised with only the labeled data
  \item \textbf{Mean-Teacher (MT)}: \cite{Tarvainen2017MeanTeacher} adapted for BEV segmentation (base model is LSS)
  \item \textbf{UniMatch}: \cite{Yang2022UniMatch} adapted for BEV segmentation with our proposed CamDrop and BFD (base model is LSS)
\end{itemize}
Note that SSL for multi-camera BEV segmentation is under-explored and there are no existing methods for comparison.
We also apply CamDrop and BFD to MT to show the effectiveness of our proposed input and feature perturbations.

In Table \ref{tab:ssl_nusc}, we show the results of our SSL experiments on the nuScenes dataset.
Compared to the supervised only method, our proposed PCT+MT method outperforms the baseline by a significant margin for all SSL splits.
It can also be seen that PCT alone can perform better than supervised only baseline and can even perform competitively against SSL methods like MT and UniMatch.
PCT, when combined with MT and strong perturbations (CamDrop and BFD), achieves the best performance for all SSL splits, improving on the difficult $1/16$ split by a significant margin of $7.7\%$.
We obtain similar gains with PCT+MT with CamDrop and BFD on the Argoverse2 dataset, as shown in \refappendix{apdx:av2}.

CamDrop and BFD have been shown to be effective in improving the performances of baseline and proposed method.
MT with CamDrop improves MT on most splits, and MT with BFD improves MT on all splits.
As shown in Figure \ref{fig:qualitative}, MT with CamDrop and BFD improves the performance of MT for the difficult $1/16$ split, especially for areas further away.
The BEV segmentation predictions for PCT+MT with CamDrop and BFD, show improvements for dividers and smaller categories like ``Stop Line.''

\subsection{Unsupervised Domain Adaptation Results}
\label{sec:uda_results}

\begin{table}[h]
\centering
\caption{Unsupervised Domain Adaptation results on the nuScenes dataset. We experimented with four different domain gaps and evaluated on the validation set. The results are in IoU (\%) and the best results are in \textbf{bold}.}
\label{tab:da_nusc}
\footnotesize
\setlength{\tabcolsep}{4pt}
\begin{tabular}{c||c|c|c|c|c|c|c}
\hline
Method & Drive. & Cross. & Walk. & Stop. & Car. & Div. & Mean  \\
\hline
\multicolumn{8}{c}{\textbf{Day $\rightarrow$ Night}} \\
\hline
Baseline    & 30.5 & 1.7  & 4.0  & 1.9  & 0    & 11.8 & 8.3 \\
DomainAdv   & 47.1 & 16.1 & 10.7 & 5.7  & 0    & 11.2 & 15.1 \\
FDA+MT      & 44.9 & 7.8  & 12.3 & 4.6  & 0    & 14.1 & 14.0 \\
\hdashline
PCT         & 51.3 & 19.4 & \textbf{16.1} & \textbf{7.6}  & 0    & 19.3 & 19.0 \\
PCT+CamDrop & \textbf{52.5} & \textbf{19.8} & 15.8 & 6.8  & 0    & \textbf{20.6} & \textbf{19.2} \\
\hline\hline
\multicolumn{8}{c}{\textbf{Dry $\rightarrow$ Rain}} \\
\hline
Baseline    & 74.2 & 38.2 & 46.8 & 31.3 & 39.6 & 32.7 & 43.8 \\
DomainAdv   & 72.0 & 39.8 & 42.0 & 33.7 & 38.9 & 33.6 & 43.3 \\
FDA+MT      & 75.3 & 42.0 & 47.0 & 35.3 & 39.5 & 34.2 & 45.6 \\
\hdashline
PCT         & \textbf{78.3} & \textbf{45.2} & 52.1 & \textbf{37.6} & 47.2 & 36.4 & 49.5 \\
PCT+CamDrop & \textbf{78.3} & 44.7 & \textbf{52.6} & 37.2 & \textbf{48.7} & \textbf{37.3} & \textbf{49.8} \\
\hline\hline
\multicolumn{8}{c}{\textbf{Singapore $\rightarrow$ Boston}} \\
\hline
Baseline    & 39.5 &  3.1 & 12.8 &  4.1 &  1.0 & 12.1 & 12.1 \\
DomainAdv   & 35.7 &  4.2 & 11.3 &  4.8 &  0.6 &  9.7 & 11.1 \\
FDA+MT      & 41.8 &  6.5 & 14.5 &  7.1 &  1.1 & 10.8 & 13.6 \\
\hdashline
PCT         & 47.0 &  8.0 & 19.3 &  6.3 &  0.7 & 13.7 & 15.8 \\
PCT+CamDrop & \textbf{48.9} &  \textbf{8.9} & \textbf{21.6} &  \textbf{7.6} &  \textbf{1.7} & \textbf{15.6} & \textbf{17.4} \\
\hline\hline
\multicolumn{8}{c}{\textbf{Boston $\rightarrow$ Singapore}} \\
\hline
Baseline    & 37.1 &  7.5 &  9.4 &  4.6 &  4.4 & 11.8 & 12.5 \\
DomainAdv   & 40.0 &  8.3 & 11.7 &  4.8 &  2.2 & 11.6 & 13.1 \\
FDA+MT      & 38.9 &  8.3 & 12.3 &  5.2 &  2.0 & 11.6 & 13.1 \\
\hdashline
PCT         & 46.2 &  8.6 & \textbf{14.2} &  6.4 &  3.7 & 15.0 & 15.7 \\
PCT+CamDrop & \textbf{47.2} &  \textbf{9.0} & 14.1 &  \textbf{7.7} &  \textbf{4.8} & \textbf{16.4} & \textbf{16.5} \\
\hline
\end{tabular}
\end{table}

In our UDA benchmark, we compare our proposed method against the following baselines:
\begin{itemize}
  \item \textbf{Baseline}: LSS supervised with only the source domain
  \item \textbf{DomainAdv}: Adversarial baseline used in \cite{Man2023DualCross}, which adds domain classifiers to the Image Encoder and BEV Encoder and trained with GRL \cite{Ganin2015DomainAdversarial} (base model is LSS)
  \item \textbf{FDA+MT}: Fourier Domain Adaptation with Mean-Teacher \cite{Yang2020FDA} (base model is LSS)
\end{itemize}

In Table \ref{tab:da_nusc}, we show the results of our UDA experiments on the nuScenes dataset.
The FDA+MT baseline improves upon the baseline LSS and DomainAdv in most domain gaps.
Although DomainAdv works well in the difficult Day $\rightarrow$ Night and Boston $\rightarrow$ Singapore domain gaps, it fails to perform well on others.
We believe this is due to the domain classifier not functioning correctly for the other domain gaps.

For all domain gaps, both of our proposed PCT and PCT+CamDrop significantly outperform baselines, especially for major categories like ``Drivable Area'' and ``Walkway.''
As we show in Figure \ref{fig:qualitative}, PCT has a clearer BEV segmentation for the difficult Day $\rightarrow$ Night domain gap than the rest of the baselines, as there are fewer segmentation artifacts.
The addition of CamDrop helps improve PCT in all domain gaps.
We also show the efficacy of PCT on the Argoverse2 dataset in \refappendix{apdx:av2}.

\subsection{Ablation Study}
\label{sec:ablation_studies}

\begin{table}[h]
\centering
\caption{We evaluated pseudo-labels generated from different semantic segmentation architectures. The segmentation models are trained on the Cityscapes dataset and the ``Segmentation Quality'' is mIoU of the Cityscapes validation split. The experiments for SSL and UDA are conducted with 1/16 labeled split and Day $\rightarrow$ Night domain gap, respectively.}
\label{tab:pl_quality}
\footnotesize
\begin{tabular}{c|c||c|c}
\hline
Pseudo-Labeling & Seg. Quality & SSL  & UDA \\
Model           & (mIoU)       & \textbf{1/16} & \textbf{Day $\rightarrow$ Night} \\
\hline
DeepLabV3+  & 79.5 & 23.3 & 11.7\\
Segformer   & 82.3 & 23.2 & 15.6\\
Mask2Former & 83.7 & 23.8 & 17.2\\
\hline
\end{tabular}
\end{table}

\begin{table}[h]
\centering
\parbox{0.57\linewidth}{
\centering
\caption{We evaluated different datasets for training semantic segmentation models for generating pseudo-labels. Cityscapes \cite{Cordts2016Cityscapes}, BDD100k \cite{Yu2020BDD100k}, and Synthia \cite{Ros2016SYNTHIA} are all perspective view semantic segmentation dataset. We also report the relative depth pseudo-label generated from Depth-Anything \cite{Yang2024DepthAnything}.}
\label{tab:pl_dataset}
\footnotesize
\begin{tabular}{cc||c|c}
\hline
Pseudo-Label & Training Dataset & SSL & UDA \\
Task         & (or Foundation Model) & \textbf{1/16} & \textbf{Day $\rightarrow$ Night} \\
\hline
\multirow{4}{*}{Sem. Seg.} & Cityscapes & 23.8 & 17.2\\
                           & BDD100k & 24.1 & 19.0 \\
                           & Cityscapes+BDD100k  & 24.1 & 18.4\\
                           & BDD100k+Synthia     & 23.8 & 17.6\\
\hdashline
Rel. Depth                 & Depth Anything & 22.8 & 19.0\\
\hline
\end{tabular}
}%
\hfill
\parbox{0.4\linewidth}{
\centering
\caption{Different crop sizes for training with PCT.}
\label{tab:crop_size}
\footnotesize
\begin{tabular}{c||c|c}
\hline
\multirow{2}{*}{Crop Size} & SSL  & UDA \\
                           & \textbf{1/16} & \textbf{Day $\rightarrow$ Night} \\
\hline
$128\times352$ & 20.2 & 16.9 \\
$224\times480$ & 24.1 & 19.0 \\
$360\times720$ & 24.3 & 19.8 \\
\hline
\end{tabular}
}
\end{table}

\noindent\textbf{Effect of Pseudo-Label Quality.}
Table \ref{tab:pl_quality} evaluates pseudo-labels generated from different semantic segmentation architectures trained on the Cityscapes dataset.
Higher-quality pseudo-labels result in better performance for UDA, but the difference is not as significant for SSL.
Although not experimented with, carefully annotated PV images may further improve the capability of the PCT.

\noindent\textbf{Effect of Pseudo-Label Dataset.}
In Table \ref{tab:pl_dataset}, we evaluated different datasets for training semantic segmentation models for generating pseudo-labels.
Training with pseudo-labels generated from BDD100k, known for containing various domain variations, performed the best in SSL and UDA.
While we hoped for joint datasets (e.g. Cityscapes+BDD100k) to further improve the performance, the results are similar to BDD100k alone.
PCT with relative depth pseudo-labels does not perform as well for SSL, but performs comparable to semantic segmentation with BDD100k for UDA.
With that being said, the performance disparities between the training datasets and tasks are relatively minor, and the choice of dataset and pseudo-labeling task is flexible for PCT.

\noindent\textbf{Effect of Crop Size for PCT.}
In Table \ref{tab:crop_size}, we evaluated different crop sizes for training with PCT.
In our PV Head, the resolution depends on the crop size, and the results show that the higher crop size results in better performance for both SSL and DA.
Lower crop size results in lower resolution and less context for the semantic segmentation model to learn to produce robust feature maps.

\begin{table}[h]
\centering
\parbox{0.57\textwidth}{
  \centering
  \caption{Effect of PCT on different BEV architectures.}
  \label{tab:bev_arch}
  \footnotesize
  \begin{tabular}{c|c||c|c}
  \hline
  \multirow{2}{*}{BEV Architecture} & \multirow{2}{*}{PCT} & SSL  & UDA \\
                                    &                      & \textbf{1/16} & \textbf{Day $\rightarrow$ Night} \\
  \hline
  \multirow{2}{*}{LSS \cite{Philion2020LSS}} &        & 21.9 & 8.3 \\
                       & \cmark & \textbf{24.1} & \textbf{19.0}\\
  \hdashline
  \multirow{2}{*}{CVT \cite{Zhou2022CVT}} &        & 14.4 & 9.1 \\
                       & \cmark & \textbf{16.3} & \textbf{15.2}\\
  \hdashline
  \multirow{2}{*}{BEVFusion \cite{Liu2022BEVFusion}} &        & 21.4 & 7.1 \\
                             & \cmark & \textbf{23.0} & \textbf{19.1}\\
  \hdashline
  \multirow{2}{*}{GKT \cite{Chen2022GKT}} &        & 15.5 & 10.2 \\
                       & \cmark & \textbf{16.5} & \textbf{14.3} \\
  \hline
  \end{tabular}
}%
\hfill
\parbox{0.4\textwidth}{
  \centering
  \caption{Effect of varying number of cameras to drop for CamDrop. The \# Max Drops refers to the maximum number of cameras which can be randomly dropped at once.}
  \label{tab:camdrop}
  \begin{tabular}{c||c|c}
  \hline
  \multirow{2}{*}{\# Max Drops} & SSL (MT) & UDA (PCT)\\
                                & \textbf{1/16} & \textbf{Day $\rightarrow$ Night} \\
  \hline
  0 & 23.7 & 19.0\\
  1 & \textbf{25.5} & \textbf{19.2}\\
  2 & 25.1 & 18.8\\
  3 & 25.1 & 18.3\\
  4 & 24.9 & 18.3\\
  5 & 22.4 & 18.2\\
  \hline
  \end{tabular}
}
\end{table}

\noindent\textbf{Effect of PCT on Different BEV Architectures.}
We show the effect of PCT on different BEV architectures in Table \ref{tab:bev_arch}.
The image encoder and hyperparameters are consistent across different BEV architectures.
Notably, all the BEV architectures benefit from PCT, and the performance improvements are consistent across different BEV architectures.
The results show that our proposed PCT framework is flexible and boosts the base model performances.
Additionally, PCT does not increase the network parameters in any way during inference time, as we can freely remove the PV Head.

\noindent\textbf{Effect of maximum camera drops.}
Table \ref{tab:camdrop} shows the effect of varying number of cameras to drop for CamDrop.
For both UDA and SSL, dropping one camera results in significant performance gains compared to no drop, and further dropping results in worse returns.
For methods which does not utilize MT, such as PCT, the CamDrop may be too strong when dropping more than one camera.

\begin{table}[h]
\centering
\caption{We compare our method against DualCross. ``Crop Size'' is the input PV image size. ``Modal'' refers to the modalities used for training where \textbf{C} refers to camera and \textbf{L} refers to LiDAR. The results are in IoU (\%) and the best results are in \textbf{bold}.}
\label{tab:abl_dualcross}
\footnotesize
\begin{tabular}{c|c|c||c|c|c}
\hline
Method & Crop Size & Modal & Road & Lane & Vehicle \\
\hline
\multicolumn{6}{c}{\textbf{Day $\rightarrow$ Night}} \\
\hline
DualCross                    & $128\times352$ & C+L & 51.8 & 16.9 & 17.0 \\
\hdashline
\multirow{2}{*}{PCT+CamDrop} & $128\times352$ & C & 49.5 & 17.8 & 18.3 \\
                             & $224\times480$ & C & \textbf{56.3} & \textbf{21.2} & \textbf{22.3} \\
\hline\hline
\multicolumn{6}{c}{\textbf{Dry $\rightarrow$ Rain}} \\
\hline
DualCross                    & $128\times352$ & C+L & 71.9 & 19.5 & 29.6 \\
\hdashline
\multirow{2}{*}{PCT+CamDrop} & $128\times352$ & C & 76.3 & 32.8 & 27.2 \\
                             & $224\times480$ & C & \textbf{79.3} & \textbf{36.0} & \textbf{31.5} \\
\hline\hline
\multicolumn{6}{c}{\textbf{Boston $\rightarrow$ Singapore}} \\
\hline
DualCross                    & $128\times352$ & C+L & 43.1 & \textbf{15.6} & 20.5 \\
\hdashline
\multirow{2}{*}{PCT+CamDrop} & $128\times352$ & C & 44.0 & 13.0 & 19.7 \\
                             & $224\times480$ & C & \textbf{47.8} & \textbf{15.6} & \textbf{21.8} \\
\hline
\end{tabular}
\end{table}

\subsection{Comparisons with Other UDA BEV Methods}
\label{sec:compare_dualcross}

Here, we compare our method against DualCross \cite{Man2023DualCross}, which utilizes cross-modality information with a LiDAR teacher.
The experimental setup used for DualCross is drastically different, but we made our network and hyperparameters consistent with the ones used in DualCross for a fair comparison.
More specifically, they use a modified LSS architecture with Efficient-b0.
The results are shown in Table \ref{tab:abl_dualcross}.
DualCross uses a smaller crop size of $128\times352$, and our method can perform comparable to their method without needing multi-modal information.
However, our method can perform superior to DualCross when utilizing a larger crop size of $224\times480$ since the PV Head greatly benefits from a larger crop size.
DA-BEV \cite{Jiang2024DABEV} is another UDA method, but its implementation details are unclear (e.g., UDA splits), and their code is not publicly available at the time of writing.

\section{Conclusion}
\label{sec:conclusions}

This work presents the Perspective Cue Training (PCT) framework, a novel training framework that effectively leverages unlabeled perspective view (PV) images through multi-tasking with PV tasks to address the challenges of limited BEV annotations and domain shifts.
Our experiments showed that PCT with pseudo-labels generated from publicly available models for semantic segmentation and depth estimation shows promising results for both semi-supervised learning (SSL) and unsupervised domain adaptation (UDA).
Our approach is flexible and applicable to various existing BEV architectures and demonstrates significant improvements over baseline methods.
The introduction of Camera Dropout (CamDrop) and BEV Feature Dropout (BFD) further enhances the performance of our method.

\begin{appendix}

\section{Results for Argoverse2}
\label{apdx:av2}

%

\begin{table}[h]
  \centering
  \parbox{.5\textwidth}{
    \centering
    \caption{Semi-Supervised Learning results on the Argoverse2 dataset. The results are in mIoU (\%). The best results are in \textbf{bold}.}
    \label{tab:ssl_av2}
    \footnotesize
    \begin{tabular}{c||c|c|c|c}
    \hline
    Method             & 1/16 & 1/8  & 1/4  & 1/2  \\
    \hline
    Sup. Only          & 35.9 & 42.6 & 48.1 & 53.2 \\
    PCT+MT+CamDrop+BFD & \textbf{40.9} & \textbf{45.6} & \textbf{51.2} & \textbf{54.2} \\
    \hline
    \end{tabular}
  }%
  \hfill
  \parbox{.45\textwidth}{
    \centering
    \caption{Unsupervised Domain Adaptation results on the Argoverse2 dataset. We experimented with two different ``City-to-City'' domain gaps and evaluated on the validation set. The results are in IoU (\%) and the best results are in \textbf{bold}.}
    \label{tab:da_av2}
    \footnotesize
    \setlength{\tabcolsep}{4pt}
    \begin{tabular}{c||c|c|c|c}
    \hline
    Method & Drivable. & Ped. Cross. & Divider & Mean  \\
    \hline
    \multicolumn{5}{c}{\textbf{Palo Alto $\rightarrow$ Miami}} \\
    \hline
    Baseline    & 50.1 & 9.5  & 17.7 & 25.8 \\
    PCT         & \textbf{54.9} & \textbf{12.6} & \textbf{19.3} & \textbf{28.9} \\
    \hline\hline
    \multicolumn{5}{c}{\textbf{Austin $\rightarrow$ Pittsburgh}} \\
    \hline
    Baseline    & 47.2 & 7.3  & 27.2 & 27.2 \\
    PCT         & \textbf{51.0} & \textbf{13.8} & \textbf{27.6} & \textbf{30.8} \\
    \hline
    \end{tabular}
  }
\end{table}

We provide additional results for SSL and UDA on the Argoverse2 \cite{Wilson2023Argoverse2} dataset.
We utilize static categories, including Drivable Area, Pedestrian Crossing, and Divider.
LSS is used as the base model and the hyperparameters across the experiments are consistent with our other experiments.

For SSL, we experimented with four levels of labeled data splits, similar to our nuScenes experiment, and evaluated on a common validation set.
The SSL results are provided in Table \ref{tab:ssl_av2}.
Similar to our nuScenes results, our method outperforms the supervised only baseline for all SSL splits.

For UDA, we experimented with two different ``City-to-City'' domain gaps: Palo Alto $\rightarrow$ Miami and Austin $\rightarrow$ Pittsburgh.
The UDA results are provided in Table \ref{tab:da_av2}.
The results show that our method outperforms the baseline for both domain gaps, demonstrating the effectiveness of our method for UDA on another dataset.

\end{appendix}

\bibliographystyle{unsrtnat}
\bibliography{references}  

\begin{thebibliography}{44}
\providecommand{\natexlab}[1]{#1}
\providecommand{\url}[1]{\texttt{#1}}
\expandafter\ifx\csname urlstyle\endcsname\relax
  \providecommand{\doi}[1]{doi: #1}\else
  \providecommand{\doi}{doi: \begingroup \urlstyle{rm}\Url}\fi

\bibitem[Chen et~al.(2023)Chen, Zhang, et~al.]{Chen2023VMA}
Shaoyu Chen, Yunchi Zhang, et~al.
\newblock Vma: Divide-and-conquer vectorized map annotation system for large-scale driving scene.
\newblock \emph{arxiv preprint arXiv:2304.09807}, 2023.

\bibitem[Zhang et~al.(2023)Zhang, Chen, et~al.]{Zhang2023CAMA}
Jiaxin Zhang, Shiyuan Chen, et~al.
\newblock A vision-centric approach for static map element annotation.
\newblock \emph{arxiv preprint arXiv:2309.11754}, 2023.

\bibitem[Caesar et~al.(2019)Caesar, Bankiti, et~al.]{Caesar2019nuScenes}
Holger Caesar, Varun Bankiti, et~al.
\newblock nuscenes: A multimodal dataset for autonomous driving.
\newblock In \emph{CVPR}, 2019.

\bibitem[Wilson et~al.(2023)Wilson, Qi, et~al.]{Wilson2023Argoverse2}
Benjamin Wilson, William Qi, et~al.
\newblock Argoverse 2: Next generation datasets for self-driving perception and forecasting.
\newblock \emph{arxiv preprint arXiv:2301.00493}, 2023.

\bibitem[Cordts et~al.(2016)Cordts, Omran, et~al.]{Cordts2016Cityscapes}
Marius Cordts, Mohamed Omran, et~al.
\newblock The cityscapes dataset for semantic urban scene understanding.
\newblock In \emph{CVPR}, 2016.

\bibitem[Yu et~al.(2020)Yu, Chen, et~al.]{Yu2020BDD100k}
Fisher Yu, Haofeng Chen, et~al.
\newblock Bdd100k: A diverse driving dataset for heterogeneous multitask learning.
\newblock In \emph{CVPR}, 2020.

\bibitem[Ros et~al.(2016)]{Ros2016SYNTHIA}
Germ{\'a}n Ros et~al.
\newblock The synthia dataset: A large collection of synthetic images for semantic segmentation of urban scenes.
\newblock In \emph{CVPR}, 2016.

\bibitem[Chen et~al.(2018)Chen, Zhu, et~al.]{Chen2018DLV3}
Liang-Chieh Chen, Yukun Zhu, et~al.
\newblock Encoder-decoder with atrous separable convolution for semantic image segmentation.
\newblock In \emph{ECCV}, 2018.

\bibitem[Xie et~al.(2021)Xie, Wang, et~al.]{Xie2021SegFormer}
Enze Xie, Wenhai Wang, et~al.
\newblock Segformer: Simple and efficient design for semantic segmentation with transformers.
\newblock In \emph{NeurIPS}, 2021.

\bibitem[Cheng et~al.(2021)Cheng, Misra, et~al.]{Cheng2021Mask2Former}
Bowen Cheng, Ishan Misra, et~al.
\newblock Masked-attention mask transformer for universal image segmentation.
\newblock In \emph{CVPR}, 2021.

\bibitem[Contributors(2020)]{mmseg2020}
MMSegmentation Contributors.
\newblock {MMSegmentation}: Openmmlab semantic segmentation toolbox and benchmark.
\newblock \url{https://github.com/open-mmlab/mmsegmentation}, 2020.

\bibitem[Man et~al.(2023)Man, Gui, et~al.]{Man2023DualCross}
Yunze Man, Liangyan Gui, et~al.
\newblock Dualcross: Cross-modality cross-domain adaptation for monocular bev perception.
\newblock In \emph{IROS}, 2023.

\bibitem[Philion and Fidler(2020)]{Philion2020LSS}
Jonah Philion and Sanja Fidler.
\newblock Lift, splat, shoot: Encoding images from arbitrary camera rigs by implicitly unprojecting to 3d.
\newblock In \emph{ECCV}, 2020.

\bibitem[Li et~al.(2022)Li, Ge, et~al.]{Li2022BEVDepth}
Yinhao Li, Zheng Ge, et~al.
\newblock Bevdepth: Acquisition of reliable depth for multi-view 3d object detection.
\newblock In \emph{AAAI}, 2022.

\bibitem[Liu et~al.(2023)Liu, Tang, et~al.]{Liu2022BEVFusion}
Zhijian Liu, Haotian Tang, et~al.
\newblock Bevfusion: Multi-task multi-sensor fusion with unified bird's-eye view representation.
\newblock In \emph{ICRA}, 2023.

\bibitem[Zhou and Krahenbuhl(2022)]{Zhou2022CVT}
Brady Zhou and Philipp Krahenbuhl.
\newblock Cross-view transformers for real-time map-view semantic segmentation.
\newblock In \emph{CVPR}, 2022.

\bibitem[Chen et~al.(2022)Chen, Cheng, et~al.]{Chen2022GKT}
Shaoyu Chen, Tianheng Cheng, et~al.
\newblock Efficient and robust 2d-to-bev representation learning via geometry-guided kernel transformer.
\newblock \emph{arxiv preprint arXiv:2206.04584}, 2022.

\bibitem[Klinghoffer et~al.(2023)Klinghoffer, Philion, et~al.]{Klinghoffer2023TowardsVR}
Tzofi Klinghoffer, Jonah Philion, et~al.
\newblock Towards viewpoint robustness in bird’s eye view segmentation.
\newblock In \emph{ICCV}, 2023.

\bibitem[Swerdlow et~al.(2023)Swerdlow, Xu, et~al.]{Swerdlow2023BEVGen}
Alexander Swerdlow, Runsheng Xu, et~al.
\newblock Street-view image generation from a bird's-eye view layout.
\newblock \emph{arxiv preprint arXiv:2301.04634}, 2023.

\bibitem[Borse et~al.(2023)Borse, Klingner, et~al.]{Borse2022XAlign}
Shubhankar Borse, Marvin Klingner, et~al.
\newblock X-align: Cross-modal cross-view alignment for bird's-eye-view segmentation.
\newblock In \emph{WACV}, 2023.

\bibitem[Tarvainen and Valpola(2017)]{Tarvainen2017MeanTeacher}
Antti Tarvainen and Harri Valpola.
\newblock Mean teachers are better role models: Weight-averaged consistency targets improve semi-supervised deep learning results.
\newblock In \emph{NeurIPS}, 2017.

\bibitem[Ke et~al.(2019)Ke, Wang, et~al.]{Ke2019DualSB}
Zhanghan Ke, Daoye Wang, et~al.
\newblock Dual student: Breaking the limits of the teacher in semi-supervised learning.
\newblock In \emph{ICCV}, 2019.

\bibitem[Chen et~al.(2020)Chen, Zhu, et~al.]{Chen2020AMM}
Zhihao Chen, Lei Zhu, et~al.
\newblock A multi-task mean teacher for semi-supervised shadow detection.
\newblock In \emph{CVPR}, 2020.

\bibitem[Xu et~al.(2021)Xu, Zhang, et~al.]{Xu2021SoftTeacher}
Mengde Xu, Zheng Zhang, et~al.
\newblock End-to-end semi-supervised object detection with soft teacher.
\newblock In \emph{ICCV}, 2021.

\bibitem[Devries and Taylor(2017)]{Devries2017Cutout}
Terrance Devries and Graham~W. Taylor.
\newblock Improved regularization of convolutional neural networks with cutout.
\newblock \emph{arxiv preprint arXiv:1708.04552}, 2017.

\bibitem[Yun et~al.(2019)Yun, Han, et~al.]{Yun2019CutMix}
Sangdoo Yun, Dongyoon Han, et~al.
\newblock Cutmix: Regularization strategy to train strong classifiers with localizable features.
\newblock In \emph{ICCV}, 2019.

\bibitem[French et~al.(2019)French, Laine, et~al.]{French2019SemisupervisedSS}
Geoff French, Samuli Laine, et~al.
\newblock Semi-supervised semantic segmentation needs strong, varied perturbations.
\newblock In \emph{BMVC}, 2019.

\bibitem[Yang et~al.(2022)Yang, Qi, Litong, et~al.]{Yang2022UniMatch}
Lihe Yang, Lei Qi, Litong, et~al.
\newblock Revisiting weak-to-strong consistency in semi-supervised semantic segmentation.
\newblock In \emph{CVPR}, 2022.

\bibitem[Sohn et~al.(2020)Sohn, Berthelot, et~al.]{Sohn2020FixMatch}
Kihyuk Sohn, David Berthelot, et~al.
\newblock Fixmatch: Simplifying semi-supervised learning with consistency and confidence.
\newblock In \emph{NeurIPS}, 2020.

\bibitem[Ouali et~al.(2020)Ouali, Hudelot, et~al.]{Ouali2020CCT}
Yassine Ouali, C{\'e}line Hudelot, et~al.
\newblock Semi-supervised semantic segmentation with cross-consistency training.
\newblock In \emph{CVPR}, 2020.

\bibitem[Gosala et~al.(2023)Gosala, Petek, et~al.]{Gosala2023SkyEye}
Nikhil Gosala, K{\"u}rsat Petek, et~al.
\newblock Skyeye: Self-supervised bird's-eye-view semantic mapping using monocular frontal view images.
\newblock In \emph{CVPR}, 2023.

\bibitem[Zhu et~al.(2023)Zhu, Liu, et~al.]{Zhu2023SSLMonoBEV}
Junyu Zhu, Lina Liu, et~al.
\newblock Semi-supervised learning for visual bird's eye view semantic segmentation.
\newblock \emph{arxiv preprint arXiv:2308.14525}, 2023.

\bibitem[Tzeng et~al.(2017)Tzeng, Hoffman, et~al.]{Tzeng2017ADDA}
Eric Tzeng, Judy Hoffman, et~al.
\newblock Adversarial discriminative domain adaptation.
\newblock In \emph{CVPR}, 2017.

\bibitem[Saito et~al.(2017)Saito, Watanabe, et~al.]{Saito2017MaximumCD}
Kuniaki Saito, Kohei Watanabe, et~al.
\newblock Maximum classifier discrepancy for unsupervised domain adaptation.
\newblock In \emph{CVPR}, 2017.

\bibitem[Long et~al.(2015)Long, Cao, et~al.]{Long2015LearningTF}
Mingsheng Long, Yue Cao, et~al.
\newblock Learning transferable features with deep adaptation networks.
\newblock In \emph{ICML}, 2015.

\bibitem[Yang and Soatto(2020)]{Yang2020FDA}
Yanchao Yang and Stefano Soatto.
\newblock Fda: Fourier domain adaptation for semantic segmentation.
\newblock In \emph{CVPR}, 2020.

\bibitem[Ganin et~al.(2015)Ganin, Ustinova, et~al.]{Ganin2015DomainAdversarial}
Yaroslav Ganin, E.~Ustinova, et~al.
\newblock Domain-adversarial training of neural networks.
\newblock \emph{JMLR}, 2015.

\bibitem[Hoyer et~al.(2021)Hoyer, Dai, et~al.]{Hoyer2021DAFormer}
Lukas Hoyer, Dengxin Dai, et~al.
\newblock Daformer: Improving network architectures and training strategies for domain-adaptive semantic segmentation.
\newblock In \emph{CVPR}, 2021.

\bibitem[Zhao et~al.(2023)Zhao, Mithun, et~al.]{Zhao2023UnsupervisedDA}
Xingchen Zhao, Niluthpol~Chowdhury Mithun, et~al.
\newblock Unsupervised domain adaptation for semantic segmentation with pseudo label self-refinement.
\newblock \emph{arxiv preprint arXiv:2310.16979}, 2023.

\bibitem[Jiang et~al.(2024)Jiang, Huang, et~al.]{Jiang2024DABEV}
Kai Jiang, Jiaxing Huang, et~al.
\newblock Da-bev: Unsupervised domain adaptation for bird's eye view perception.
\newblock \emph{arxiv preprint arXiv:2401.08687}, 2024.

\bibitem[Yang et~al.(2024)Yang, Kang, et~al.]{Yang2024DepthAnything}
Lihe Yang, Bingyi Kang, et~al.
\newblock Depth anything: Unleashing the power of large-scale unlabeled data.
\newblock \emph{arxiv preprint arXiv:2401.10891}, 2024.

\bibitem[Xiao et~al.(2018)Xiao, Liu, et~al.]{Xiao2018UPerNet}
Tete Xiao, Yingcheng Liu, et~al.
\newblock Unified perceptual parsing for scene understanding.
\newblock In \emph{ECCV}, 2018.

\bibitem[Ben-David et~al.(2010)]{BenDavid2010UDA}
Shai Ben-David et~al.
\newblock A theory of learning from different domains.
\newblock \emph{Machine Learning}, 2010.

\bibitem[Srivastava et~al.(2014)Srivastava, Hinton, et~al.]{Srivastava2014Dropout}
Nitish Srivastava, Geoffrey~E. Hinton, et~al.
\newblock Dropout: a simple way to prevent neural networks from overfitting.
\newblock \emph{JMLR}, 2014.

\end{thebibliography}

\end{document}